\pdfoutput=1

\documentclass[11pt]{article}

\usepackage{acl}

\usepackage{times}
\usepackage{latexsym}
\usepackage{graphicx}
\usepackage{nicefrac}       
\usepackage{microtype}      
\usepackage{xcolor}         
\usepackage{booktabs}
\usepackage{color,colortbl}
\usepackage{multirow}
\usepackage[normalem]{ulem}
\usepackage{algorithm}
\usepackage{algpseudocode}
\usepackage{pifont}
\usepackage{amsmath}
\newcolumntype{P}[1]{>{\centering\arraybackslash}p{#1}}

\usepackage[T1]{fontenc}

\usepackage[utf8]{inputenc}

\usepackage{microtype}

\usepackage{inconsolata}

\newcommand{\dataset}[1]{\textsc{LoCoMo}}

\newcommand{\Sref}[1]{\S\ref{#1}}

\title{Evaluating Very Long-Term Conversational Memory of LLM Agents}

\author{Adyasha Maharana\textsuperscript{1} \;\;\;\;\;\; Dong-Ho Lee\textsuperscript{2} \;\;\;\; Sergey Tulyakov\textsuperscript{3} \\ \bf{Mohit Bansal\textsuperscript{1$\dagger$} \;\;\;\;\;\; Francesco Barbieri\textsuperscript{$\dagger$} \;\;\;\; Yuwei Fang\textsuperscript{3$\dagger$}}\vspace{10pt} \\
University of North Carolina, Chapel Hill\textsuperscript{1}\quad University of Southern California\textsuperscript{2}\quad Snap Inc.\textsuperscript{3}}

\begin{document}

\maketitle

\begin{abstract}

Existing works on long-term open-domain dialogues focus on evaluating model responses within contexts spanning no more than five chat sessions. Despite advancements in long-context large language models (LLMs) and retrieval augmented generation (RAG) techniques, their efficacy in \textit{very} long-term dialogues remains unexplored.
To address this research gap, we introduce a machine-human pipeline to generate high-quality, \textit{very} long-term dialogues by leveraging LLM-based agent architectures and grounding their dialogues on personas and temporal event graphs. Moreover, we equip each agent with the capability of sharing and reacting to images. The generated conversations are verified and edited by human annotators for long-range consistency and grounding to the event graphs. Using this pipeline, we collect \dataset{}, a dataset of \textit{very} long-term conversations, each encompassing 300 turns and 9K tokens on avg., over up to 35 sessions. Based on \dataset{}, we present a comprehensive evaluation benchmark to measure long-term memory in models, encompassing question answering, event summarization, and multi-modal dialogue generation tasks. Our experimental results indicate that LLMs exhibit challenges in understanding lengthy conversations and comprehending long-range temporal and causal dynamics within dialogues. Employing strategies like long-context LLMs or RAG can offer improvements but these models still substantially lag behind human performance.\footnote{Code and data to be available at \\ \url{https://snap-research.github.io/locomo}}
\end{abstract}

\makeatletter{\renewcommand*{\@makefnmark}{}
\footnotetext{$^\dagger$Equal advising.}
\makeatother}

\section{Introduction}
\label{sec:intro}

\begin{figure}[t]
    \centering
    \includegraphics[width=0.48\textwidth]{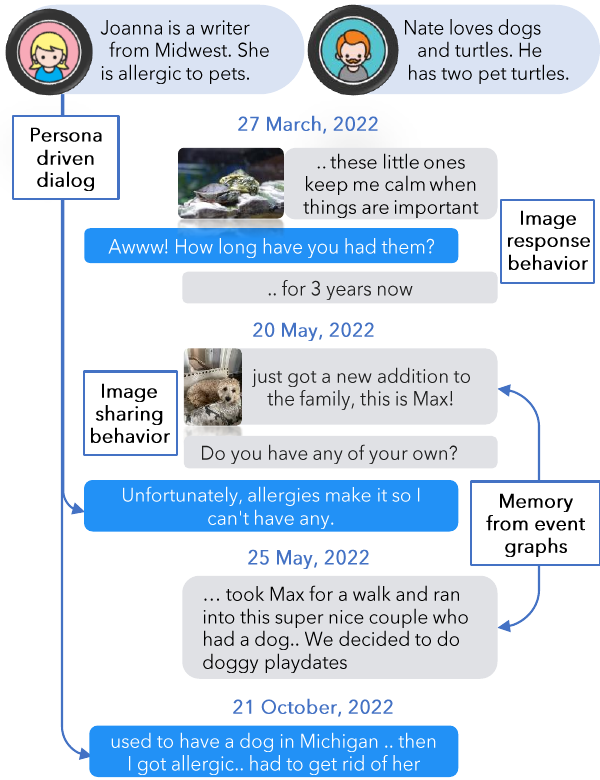}
    \caption{\textbf{An example in \dataset{}.} Dialogs are steered by the speakers' personas and corresponding events e.g., Joanna's responses are consistent with her pet allergies. For Nate, the event \textit{got a new dog} is followed by a \textit{playdate with neighbor's dog}, showcasing long-term memory. Multimodal dialog is enabled with image-sharing and image-response behaviors.
    }
    \label{fig:intro}
    \vspace{-15pt}
\end{figure}

\begin{table*}[t!]
\addtolength{\tabcolsep}{-4pt}
	\centering
	\resizebox{0.99\textwidth}{!}{
		\begin{tabular}{p{2.7in}P{0.7in}P{0.9in}P{0.9in}P{1.2in}P{0.7in}P{2.0in}}
            \toprule
            \textbf{Dataset} & \textbf{Avg. turns per conv.} & \textbf{Avg. sessions per conv.} & \textbf{Avg. tokens per conv.} & \textbf{Time Interval} & \textbf{Multimodal} &  \textbf{Collection} \\
            \midrule
            MPChat \cite{ahn2023mpchat} & 2.8 & 1 & 53.3 & - & \ding{51} & Reddit \\
            MMDialog \cite{feng2023mmdialog} & 4.6 & 1 & 72.5 & - & \ding{51} & Social media\\
            Daily Dialog \cite{li2017dailydialog} & 7.9 & 1 & 114.7 & - & \ding{55} & Crowdsourcing \\
            SODA \cite{kim-etal-2023-soda} & 7.6 & 1 & 122.4 & - & \ding{55} & LLM-generated \\
            MSC \cite{xu2022beyond} (train; 1-4 sessions) & 53.3 & 4 & 1,225.9 & few days & \ding{55} & Crowdsourcing \\
            Conversation Chronicles \cite{jang2023conversation} & 58.5 & 5 & 1,054.7 & few hours - years & \ding{55} & LLM-generated \\
            \textbf{\dataset{} (ours)} & 304.9 & 19.3 & 9,209.2 & few months & \ding{51} & LLM-gen. + crowdsourc. \\           
            \bottomrule
        \end{tabular}
	}
        \vspace{-5pt}
        \caption{\textbf{Statistics of \dataset{}} compared to existing dialog datasets. The average length of a conversation in \dataset{} is 9x that of MSC \cite{xu2022beyond}, distributed over 6x more turns and 4x more sessions (on average).}
	\label{tab:compare_datasets}
\end{table*}

\begin{figure*}[t]
    \centering
    \includegraphics[width=\textwidth]{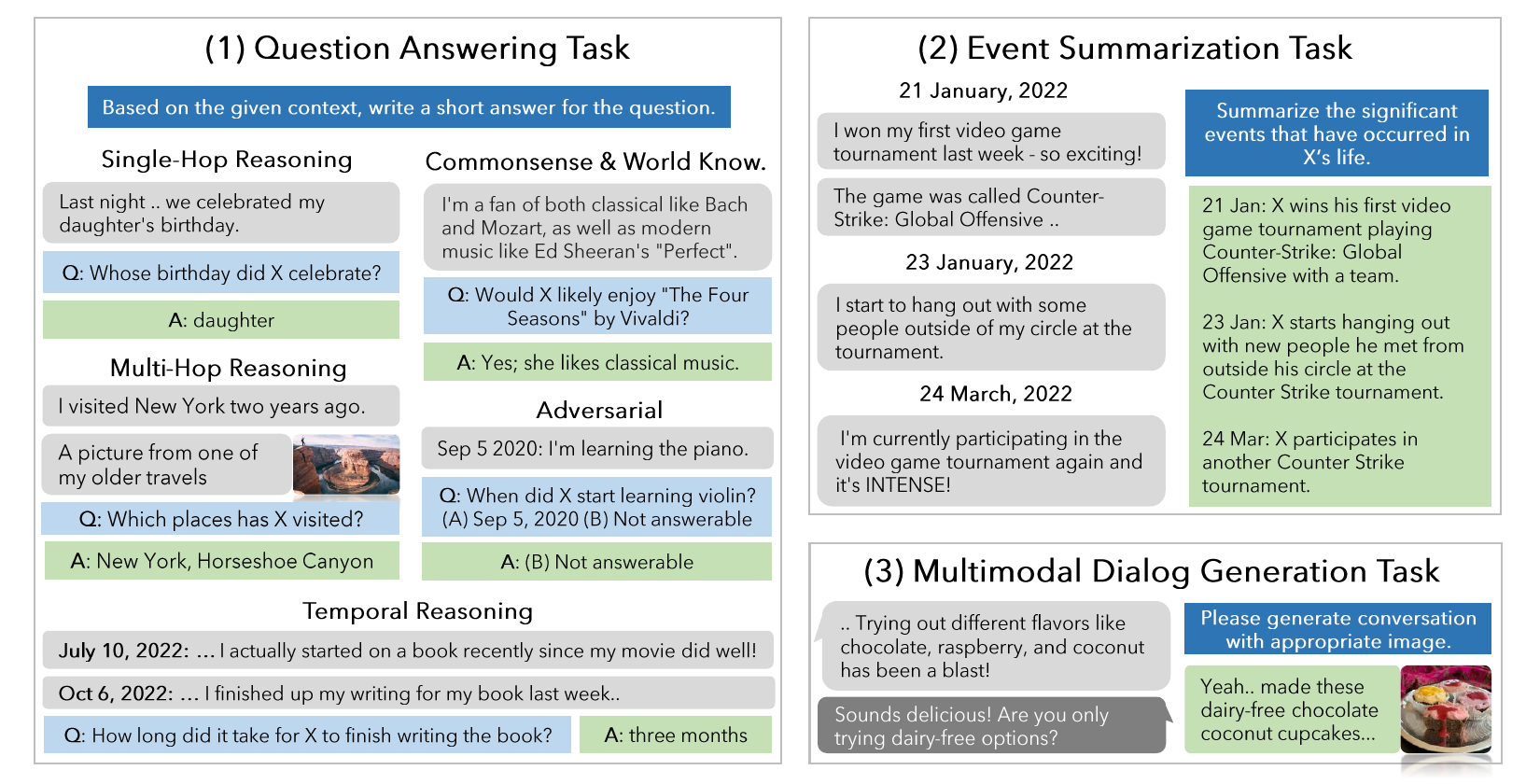}
    \vspace{-20pt}
    \caption{\textbf{Overview of our evaluation framework}. We propose three tasks: question answering, event summarization and multimodal dialog generation to evaluate models' comprehension in very long-term dialogues.}
    \label{fig:eval-framework}
    \vspace{-1pt}
\end{figure*}

Despite recent advancements in dialogue models based on LLMs for extended contexts~\cite{bertsch2024unlimiformer, xiao2023efficient}, as well as the integration of retrieval augmented generation (RAG) techniques~\cite{shuster2021retrieval, ram2023context, shi2023replug}, there is still a need for thorough evaluation of their efficacy in handling very long conversations. Indeed, studies in long-term open-domain dialogues have concentrated on assessing model responses within limited contexts e.g., $\sim$1K tokens over five chat sessions~\cite{xu2022beyond, jang2023conversation, zhang2023mind}. This long term evaluation is crucial for refining engaging chatbots capable of remembering key information from past interactions, to generate empathetic, consistent, and useful responses.

To this end, we present the first study of very long-term open-domain multi-modal dialogues, closely mirroring real-world online interactions, collected via a human-machine pipeline where we first use LLM-based generative agents to generate conversations and then ask human annotators to fix any long-term inconsistencies in the conversations.
Specifically, drawing on the understanding that real-world conversations are a complex blend of collective memories~\cite{assmann1995collective, hirst2008towards}, individual viewpoints~\cite{hirst2018collective}, external influences~\cite{hirst2012remembering}, and the unique persona of the speakers~\cite{pruitt2003personas, cooper1999inmates, zhou2020design, shum2019sketch}, we create \textit{very long-term} dialogues based on LLM agent with the following features:
(1) a unique persona (\Sref{ssec:dataset-persona});
(2) a timeline of causally interlinked events in their lives (\Sref{ssec:temporal-event}); and (3) \textit{reflect \& response} mechanism to respond based on dialogue history (like in ~\citet{park2023generative}) and \textit{image sharing \& image reaction} behavior which sends or reacts to images (\Sref{ssec:llm-agent}). Finally, human annotators fix long-range inconsistencies in dialogues, remove irrelevant images, and verify the grounding of dialogs to events (\Sref{ssec:manual-filter}). With this pipeline, we create \dataset{}, a dataset of 50 \textit{very long-term} dialogues, each consisting of 300 turns and 9K tokens on avg., over up to 35 sessions (see Figure~\ref{fig:intro} and  Table~\ref{tab:compare_datasets}).

Conventional approaches for evaluating conversational agents in open-domain dialogues involves directly evaluating the agent response based on past dialogue history.
It often employs lexical overlap~\cite{papineni-etal-2002-bleu} and semantic overlap~\cite{zhang2019bertscore} between ground truth and the agent response, or consistency~\cite{ghazarian2022deam}, contradiction~\cite{nie2021like, welleck2019dialogue}, and empathy~\cite{zhang2021dynaeval, zhang2022fined} of the agent response. However, these evaluation metrics are not well-suited for directly assessing the agent's comprehension of long-term contexts.

In this study, we present a holistic evaluation framework to assess an agent's proficiency in managing and responding within long-term contexts (see Figure~\ref{fig:eval-framework}). First, agents need to ``recall'' past context correctly to integrate relevant information into future responses. We present a direct examination of their \textit{memory} via a \textit{question answering} (QA) task (\Sref{ssec:benchmark-qa}). We classify questions into five distinct reasoning types to evaluate memory from multiple perspectives: single-hop, multi-hop, temporal, commonsense or world knowledge, and adversarial. Second, agents also need to recognize long-range causal and temporal connections in the dialogues to generate empathetic and relevant responses.
We propose a measurement of their causal and temporal understanding with an \textit{event graph summarization} task (\Sref{ssec:benchmark-event}). In this task, the event graphs linked to each LLM speaker serve as the correct answers, and models are tasked with extracting this information from the conversation history. Third, conversational agents need to utilize relevant context recalled from past conversations to generate responses that are consistent with the ongoing narrative. We assess this ability via the \textit{multi-modal dialog generation} task (\Sref{ssec:benchmark-mm}).

We present extensive experimental results on the \dataset{} benchmark using instruction-based LLMs, long-context LLMs, and RAG techniques (\Sref{sec:baselines}). 
Our findings include:

(1) Long-context LLMs and RAG demonstrate effectiveness in QA tasks, improving `memory' capabilities of LLMs (with improvements ranging from 22-66\%), but still significantly lag behind human levels (by 56\%), especially in temporal reasoning, (by 73\%); 

(2) long-context LLMs demonstrate significant difficulty with adversarial questions in the QA task, showing a performance that is 83\% lower than the base model. They are especially prone to misassigning dialogs or events to the wrong speaker. 
Moreover, they show poor performance on \textit{event graph summarization}, lagging behind the base model by 14\%, indicating that they may grasp the factual elements within the entire conversation but do not accurately comprehend the context; and 

(3) RAG offers a balanced compromise, combining the accuracy of short-context LLMs with the extensive comprehension of wide-context LLMs, and does particularly well when dialogues are transformed into a database of assertions (\textit{observations}) about each speaker's life and persona.

\section{Related Work}

\paragraph{Long-term Dialogue.} 
Recent approaches involve retrieving historical context from a range of previous dialogues and reasoning over retrieved segments in a temporal order~\cite{lee2023prompted, lu2023memochat, zhong2023memorybank, liang2023unleashing} and/or using events to scaffold the dialogues \cite{jang2023conversation, zhang2023mind} to enable consistency in long-term conversations. Some limitations of such frameworks are:
(1) The accuracy of retrieval can be compromised, as the retrieval model is generally trained on tasks focusing on semantic similarity rather than specifically on such dialogues.
Additionally, real-world dialogues often feature co-references and missing content (\textit{i.e.,} anaphora)~\cite{anantha2021open}, which further complicate the retrieval process~\cite{mallen-etal-2023-trust, gao2023enabling, liu2023evaluating};
(2) Challenges arise in reasoning over retrieved documents, especially when the model struggles to identify the correct context among the retrieved data~\cite{liu2023lost};
(3) Reasoning over time intervals presents challenges. For example, the way a system responds about past events can vary depending on the amount of time that has passed since the last conversation~\cite{zhang2023mind, jang2023conversation}.
Therefore, it is essential to have conversations of considerable length, as well as a systematic evaluation framework, to accurately assess the effectiveness of approaches to long-term dialogue generation. We design a long-term conversation generation pipeline based on retrieval augmentation and events graphs and propose a framework for evaluating long-term dialog agents.

\paragraph{Multi-modal Dialogue.}
Multi-modal dialogue primarily consists of two types of tasks: image-grounded dialogue and image-sharing dialogue. 
The image-grounded dialogue task is centered around responding to questions~\cite{antol2015vqa, das2017visual, kottur2019clevr} or creating natural conversations related to specific images~\cite{mostafazadeh2017image, shuster2020image, meng2020openvidial, zheng2021mmchat}. 
Conversely, the image-sharing dialogue task focuses on selecting images that semantically align with the provided dialogue context~\cite{zang2021photochat, feng2023mmdialog, lee2023dialogcc}. We use a method from the image-sharing dialogue task to create multimodal dialogs which are then evaluated as an image-grounded dialogue task.

\paragraph{Synthetic Evaluation Benchmark.}
Faced with a shortage of human-generated data and observing that LLMs are approaching the quality of human-level annotations~\cite{he2023annollm, lee-etal-2023-making}, there has been a surge in research drawing inspiration from this development. 
Consequently, numerous studies have started utilizing LLMs to augment or synthesize large-scale dialogue benchmarks for assessing responses in everyday social interactions~\cite{kim-etal-2023-soda}, examining responses in multi-modal environment~\cite{feng2023mmdialog}, and evaluating responses that align with specific persona~\cite{jandaghi2023faithful}. We leverage LLMs to create data but ensure its high quality with human verification and editing.

\section{Generative Pipeline for \dataset{}}
\label{sec:dataset}

\begin{figure*}[t]
    \centering
    \includegraphics[width=0.99\textwidth]{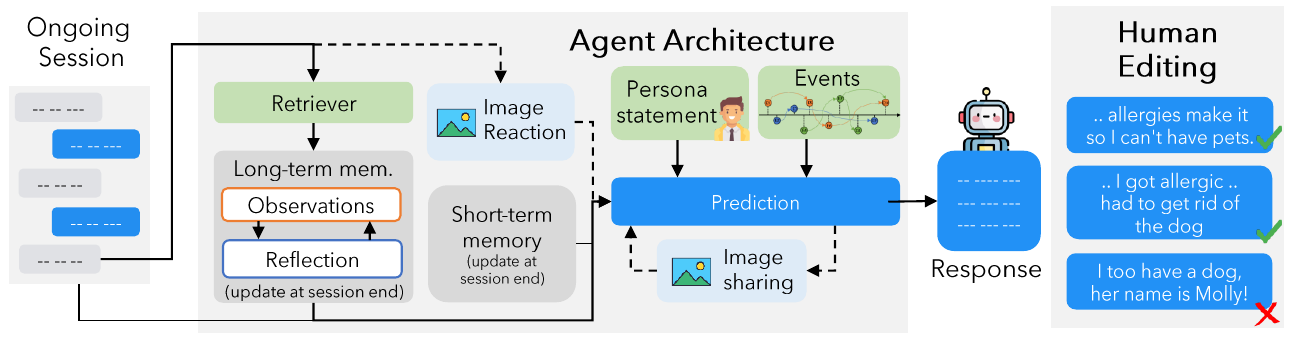}
    \vspace{-1pt}
    \caption{\textbf{Overview of the generative pipeline for \dataset{}.} Each LLM agent is assigned a distinct persona and a timeline of causally connected events in their file. The agent is equipped with a memory and reflection module to retrieve relevant history for dialog generation and is also enabled for image-sharing and image-reaction behaviors (left). The generated conversations are edited by human annotators to maintain long-range consistency (right).}
    \label{fig:main}
    \vspace{-1pt}
\end{figure*}

An overview of our generative pipeline for \dataset{} is shown in Figure~\ref{fig:main}.
We create two virtual agents, 
named $\mathcal{L}_1$ and $\mathcal{L}_2$, each initialized with a LLM $\mathcal{M}$ (\textit{i.e.,} \texttt{gpt-3.5-turbo}).
To start, unique persona statements $p$ are assigned to each agent $\mathcal{L}_i$, ensuring the integration of distinct personalities into their dialogues (\Sref{ssec:dataset-persona}).
To mirror real-life experiences, we create a temporal event graph $\mathcal{G}$ for each agent, which illustrates a realistic sequence of life events (\Sref{ssec:temporal-event}).
The LLM agent architecture~\cite{park2023generative} is utilized for each agent $\mathcal{L}_i$, enabling them to effectively memorize and reflect conversation history into ongoing dialogues (\Sref{ssec:llm-agent}).
Further, each agent $\mathcal{L}_i$ can share coherent images, thereby enhancing the multi-modal dialogue aspect.
Finally, human annotators are tasked with manually filtering and refining the generated data (\Sref{ssec:manual-filter}).

\subsection{Persona} 
\label{ssec:dataset-persona}
We select an initial persona statement $p_c$ from the MSC dataset~\cite{xu2022beyond}, encompassing 4 to 5 sentences, 
and employ \texttt{gpt-3.5-turbo} as $\mathcal{M}$ to expand these into full persona statement $p$ (See examples and prompt details in Appendix~\ref{ssec:persona_appendix}). The generated statements typically include details about one or more of the following elements~\cite{gao2023peacok}: objectives, past experiences, daily habits, and interpersonal relationships, as well as name, age, and gender of the individual.

\subsection{Temporal Event Graph} 
\label{ssec:temporal-event}
To utilize the real-life experiences of each agent in the conversation, we construct a temporal event graph, labeled as $\mathcal{G}$, for each agent. 
This graph $\mathcal{G}$, consisting of events $e_{i}$, is produced by applying the condition of $\mathcal{M}$ (i.e., \texttt{text-davinci-003}) on a designated persona $p$. Each event $e_{i}$ is associated with a date of occurrence $t_{i}$.
$\mathcal{G}$ includes causal connections $l = (e_i, e_j)$ that illustrate the causal relationships among events $e_i \in \mathcal{G}$ and reflect a natural succession of events in an individual's life. For each $\mathcal{G}$, we create up to 25 events, spread across a time frame of 6 to 12 months, in an iterative process that balances between inference time and the coherence of temporal and causal connections in the timeline. Initially, a small batch of $k=3$ events is generated, which is then used iteratively as input prompt to create the subsequent batch of $k$ events. See details in Appendix~\ref{ssec:event_appendix}.

\subsection{Virtual Agent Architecture}
\label{ssec:llm-agent}
Every agent $\mathcal{L}_i$  incorporates modules from generative agent architecture~\cite{park2023generative}.
The agent has two functions: (1) \textit{reflect \& respond}; and (2) \textit{image sharing \& image reaction}.
The agent is asked to primarily use the \textit{reflect \& respond} function while employing \textit{image sharing \& image reaction} function judiciously and appropriately within the context of the conversation.

\paragraph{Reflect \& Respond.}
The fundamental process for each agent to \textit{reflect and respond} involves the concept of short-term and long-term memory.
During inference, agent $\mathcal{L}_i$ conditions its responses on both short and long-term memories, paralleling how humans remember recent conversations while also recalling distilled important experiences from long-term memory. Following each session $k$, each agent is asked to produce a summary $w_k$ that is then stored in the short-term $\mathcal{H}_s$. This summary $w_k$ is generated by conditioning $\mathcal{M}$ on both the most recent session conversation history $h_{k}$ and the preceding summary $w_{k-1} \in \mathcal{H}_l$. For each turn $j$ within session $k$, a single turn of the conversation $h_{k_j}$ is transformed into an observation $o_{k_j}$ and then stored in the long-term memory $\mathcal{H}_l$. Then, agent $\mathcal{L}_i$ generates a response in session $k+1$ on the date $t_{k+1}^s$ by basing it on the latest summary $w_k$, reflections based on the retrieved relevant observations $o \in \mathcal{H}_s$, the ongoing conversation history in the current session $h_{k+1}$ and persona statement $p$. Long-term temporal narratives are induced in the conversation by additionally conditioning the agent's response on the subset of events in $\mathcal{G}$ that occur between the last and current session i.e. $\{e\in\mathcal{G}\,|\,t_{k}^s\,<\,t_{i}^e\,<\,t_{k+1}^s\,\}$. See details in Appendix~\ref{sssec:agent_appendix}.

\paragraph{Image Sharing \& Image Reaction.}
The \textit{image sharing \& image reaction} functions are integrated to add a multi-modal dimension to the long-term dialogues.\footnote{Image captions are also saved to long-term memory.}
The \textit{image sharing} function is called when the agent decides to send an image. This process includes:
(1) Generate a caption $c$ for the intended image using $\mathcal{M}$;
(2) Convert the caption $c$ into relevant keywords $w$ using $\mathcal{M}$;
(3) Use the keywords $k$ to find an image through web search $WEB(k)$\footnote{\url{https://pypi.org/project/icrawler/}};
(4) Share the chosen $image$.
Conversely, the \textit{image reaction} function is triggered upon receiving an image from another agent and entails:
(1) Generate caption $c$ for the received image\footnote{We use BLIP-2 \cite{li2023blip} as the captioning model.};
(2) Generate a reaction for the received image in response using $\mathcal{M}$ (See Appendix~\ref{sssec:agent_appendix}).

\subsection{Human Verification \& Editing}
\label{ssec:manual-filter}
In the concluding phase, human annotators are tasked with (1) editing the dialogue to eliminate long-term inconsistencies, (2) removing or substituting irrelevant images, and (3) verifying and editing for alignment between event graphs and the content of the conversations. Overall, we observed that annotators edited nearly 15\% of the dialog turns and removed or substituted approx. 19\% images present in the LLM-generated dataset. See examples of some edits in Appendix~\ref{ssec:edits_appendix}.

\section{\dataset{} Evaluation Benchmark}
Based on the dialogues generated in section~\ref{sec:dataset}, we introduce an evaluation benchmark (see Figure~\ref{fig:eval-framework}) composed of three tasks to assess the accuracy of \textit{long-term memory}. See statistics of the dataset and evaluation benchmark in Table~\ref{tab:dataset_statistics} in the Appendix.

\subsection{Question Answering Task}
\label{ssec:benchmark-qa}
A conversational agent is expected to possess a \textit{memory} to remember previous dialogues, reflecting it to create more engaging responses in future conversations.
For a comprehensive assessment of this \textit{memory}, we introduce a question-answering task divided into five distinct reasoning categories:
(1) \textbf{Single-hop} questions require answers based on a single session;
(2) \textbf{Multi-hop} questions require synthesizing information from multiple different sessions;
(3) \textbf{Temporal reasoning} questions can be answered through temporal reasoning and capturing time-related data cues within the conversation;
(4) \textbf{Open-domain knowledge} questions can be answered by integrating a speaker's provided information with external knowledge such as commonsense or world facts;
(5) \textbf{Adversarial} questions are designed to trick the agent into providing wrong answers, with the expectation that the agent will correctly identify them as unanswerable. 

For each category, we calculate the F1 score for exact matches, following the normalization of both the predicted and the actual ground truth answers.
However, evaluating long-form answers with automated metrics often presents challenges~\cite{xu2023critical}. LLMs tend to produce paraphrased responses in varied formats, complicating exact match evaluation. To simplify evaluation in our task, we ensure that answers in our QA annotations are directly taken from the conversations as much as possible. We instruct the LLMs to replicate the exact wording in the conversation when feasible and employ the F1 partial match metric for evaluating the predictions. Each QA sample is also annotated with the turn IDs in the conversation logs that contain the answer. We report the accuracy of retrieving the correct context for RAG models.

\subsection{Event Summarization Task}
\label{ssec:benchmark-event}
The conversation is generated based on a temporal event graph $\mathcal{G}$ which is constructed by conditioning an LLM on a persona statement $p$, reflecting the chronological sequence of events in an individual's life.
A conversational agent is expected to not only comprehend the causal connections and the sequence of events in $\mathcal{G}$ but also to recount these events as required.
To evaluate the agent's grasp of event dynamics, we introduce the event summarization task which challenges the agent to summarize the events within a designated timeframe and compares the agent's summary with events in $\mathcal{G}$. The events in \dataset{} are densely annotated lists of life events that are hard to summarize due to temporal and causal coreferences present in the dialogues, in contrast to existing summarization benchmarks of research papers \cite{li2023loogle}, movie scripts \cite{chen2022summscreen}, books \cite{kryscinski2022booksum}, emails \cite{zhang2021emailsum} etc.

Traditional metrics like BLEU~\cite{papineni-etal-2002-bleu} and ROGUE~\cite{lin-2004-rouge} focus on lexical similarity between the reference and generated summaries, not meeting our needs as we emphasize factual accuracy in summarization.
In this context, we employ FactScore~\cite{min2023factscore}, a method that evaluates the factuality of generated text by decomposing both the reference and hypothesis into atomic facts.
We adapt the metric to measure
(1) \textit{precision} of the summarized content by counting the number of atomic facts within the content that correspond with those in $\mathcal{G}$;
(2) \textit{recall} of the summarized content by determining how comprehensively the atomic facts of $\mathcal{G}$ are represented within the content.
We present the F1 score, derived from the calculated precision and recall.

\subsection{Multi-Modal Dialogue Generation Task}
\label{ssec:benchmark-mm}
The conversations in our dataset are anchored to specific personas $p$ and corresponding events $\mathcal{G}$ tailored to $p$.
The topics in conversations evolve from events that were introduced in earlier dialogues, spanning weeks or months.
This structure allows for an assessment of whether conversational agents can sustain a coherent persona and a continuous narrative over time.
For example, if a speaker recently had an injury, the next conversations would likely focus on them recuperating, rather than engaging in adventurous activities.
We assess such consistency by measuring how closely the predicted multi-modal dialogues align with the ground truth multi-modal dialogues in our dataset, quantifying this alignment through MMRelevance~\cite{feng2023mmdialog}, in addition to other NLG metrics.
\section{Experimental Setup}
\label{sec:baselines}

For the question-answering and event summarization tasks, we replace images in \dataset{} with their captions \cite{li2023blip}, and use state-of-art LLMs to reason over text-only dialogues interleaved with image captions. We use images directly for the multimodal dialog generation task only. See additional details in Appendix~\ref{sec:experiments_appendix}.

\paragraph{Question Answering.}
We evaluate three types of models:
(1) \textbf{Base} LLMs operating with constrained context lengths where earlier dialogues are omitted i.e., Mistral-7B~\cite{jiang2023mistral}, LLama-70B-chat~\cite{touvron2023llama}, \texttt{gpt-3.5-turbo}~\footnote{https://platform.openai.com/docs/models/gpt-3-5}, and \texttt{gpt-4-turbo}~\footnote{https://platform.openai.com/docs/models/gpt-4-and-gpt-4-turbo}; 
(2) \textbf{Long-context} LLMs with an extended context window i.e., \texttt{gpt-3.5-turbo-16k}; 
(3) \textbf{Retrieval-augmented Generation (RAG)} involves retrieving relevant context from a database of dialog history, observations (assertions about speakers; see \Sref{ssec:llm-agent}, Figure~\ref{fig:observation_prompt}), or session-level summaries (see \Sref{ssec:llm-agent}, Figure~\ref{fig:summary_prompt}). We employ DRAGON~\cite{lin-etal-2023-train} as retriever and \texttt{gpt-3.5-turbo-16k} as reader.

\paragraph{Event Summarization.}
We present experiments using \textbf{Base} and \textbf{Long-context} setups from the question-answering task, but refrain from including RAG since summarization requires a comprehensive understanding of the entire dialogue, rather than just retrieving a specific portion. We implement incremental summarization i.e., iteratively create a summary of a preceding sessions and then use that summary as a basis to summarize the subsequent sessions \cite{chang2023booookscore}.

\paragraph{Multi-modal Dialogue Generation.}
We generate 50 conversations using our automated pipeline (without human filtering; \Sref{sec:dataset}) for training data and train three versions of MiniGPT-5~\cite{zheng2023minigpt}:
(1) \textbf{Base} trains on prior dialogue turns only;
(2) \textbf{+ summary} trains on prior dialogue turns and a global summary of the ongoing conversation;
(3) \textbf{+ observation} trains on prior dialogue turns and observations retrieved from conversation history.
Each run is initialized with a MiniGPT-5 checkpoint finetuned on MMDialog \cite{feng2023mmdialog}.

\begin{table*}[t!]
	\centering
	\small
	\resizebox{0.99\textwidth}{!}{
		\begin{tabular}{ccccccccc}
            \toprule
            \multirow{2}{*}{\textbf{Category}} & \multirow{2}{*}{\textbf{Model}} & \multirow{2}{*}{\textbf{Context Length}} & \multicolumn{6}{c}{\textbf{Answer Prediction (F1)}} \\
            \cmidrule(lr){4-9} & & & Single Hop & Multi Hop & Temporal & Open Domain &  Adversarial & \textbf{Overall}\\
            \midrule
                \rowcolor{blue!20} Human & Human & -  & 95.1 & 85.8 & 92.6 & 75.4 & 89.4 & 87.9  \\
            \midrule
                \multirow{4}{*}{Base} & \texttt{Mistral-Instruct-7B} & 8K  & 10.2 & 12.8 & 16.1 & 19.5 & 17.0 & 13.9 \\
                & \texttt{Llama-2-Chat-70B} & 4,096  & 19.7 & 14.4 & 13.3 & 15.9 & 22.1 & 17.9 \\
                & \texttt{GPT-3.5-turbo} & 4,096  & \textbf{29.9} & 23.3 & \textbf{17.5} & \textbf{29.5} & 12.8 & 22.4 \\
                & \texttt{GPT-4-turbo} & 4,096  & 23.4 & \textbf{23.4} & 10.4 & 24.6 & \textbf{70.2} & \textbf{32.1} \\

            \midrule
                \multirow{4}{*}{Long context} & \multirow{4}{*}{\texttt{GPT-3.5-turbo-16K}} & \cellcolor{gray!25} 4K  & 31.7 & 25.4 & 16.8 & 27.6 & \textbf{13.1} & 24.1  \\
                 & & \cellcolor{gray!40} 8K  & 38.8 & 31.2 & 21.0 & 35.0 & 8.4 & 25.2  \\
                 & & \cellcolor{gray!60} 12K  & 51.1 & 40.4 & \textbf{25.0} & 36.5 & 6.4 & 33.5  \\
                 & & \cellcolor{gray!75} 16K  & \textbf{56.4} & \textbf{42.0} & 20.3 & \textbf{37.2} & 2.1 & \textbf{37.8}  \\
            \bottomrule
        \end{tabular}
	}
 \vspace{-5pt}
        \caption{\textbf{Question answering performance} of \texttt{Base} and \texttt{Long-context} models. Optimal performance is in \textbf{bold}. Results are based on F1-score for answer prediction; higher is better.}
	\label{tab:qa_results}
\end{table*}

\begin{table*}[t!]
	\centering
	\small
	\resizebox{0.99\textwidth}{!}{
		\begin{tabular}{P{0.7in}P{0.2in}P{0.4in}P{0.4in}P{0.4in}P{0.5in}P{0.4in}P{0.5in}P{0.5in}P{0.5in}P{0.5in}P{0.5in}P{0.5in}P{0.5in}}
            \toprule
              &  &  \multicolumn{6}{c}{\textbf{Answer Prediction (F1 score)}} &
             & \multicolumn{5}{c}{\textbf{Recall Accuracy (R@$k$)}} \\
            \cmidrule(lr){3-8} \cmidrule(lr){9-14} \multirow{2}{*}{\textbf{Retrieval Unit}} & \multirow{2}{*}{\textbf{top-$k$}} & Single Hop & Multi Hop & Temporal & Open \newline Domain &  Adver-\newline -sarial & \textbf{Overall} & Single Hop & Multi Hop & Temporal & Open \newline Domain &  Adver- \newline -sarial & \textbf{Overall} \\
            \midrule
            \texttt{None} & -  & 29.9 & 23.3 & 17.5 & 29.5 & 12.8 & 22.4 & - & - & - & - & - & -  \\
            \midrule
                 \multirow{4}{*}{\texttt{Dialog}} & \cellcolor{gray!30} 5 & 42.9  & 19.4  & 21.3  & 35.8  & 31.9  & 31.7  & 
                 66.2  & 34.4 &  89.2 & 38.5 & 45.7 & 58.8  \\
                 & \cellcolor{gray!45} 10 & 46.3  & 26.8  & 24.8  & 37.5  & \textbf{29.8} & 34.6  & 72.8 & 247.4 &  97.3 & 53.8 &  54.3 & 67.5 \\
                 & \cellcolor{gray!60} 25 & 48.1 & 36.1 & \textbf{26.2} & \textbf{43.4} & 23.4 & \textbf{35.8} & 87.5 & 64.1 & \textbf{97.3} &  \textbf{67.9} & 69.1 & 79.9 \\
                 & \cellcolor{gray!75} 50 & \textbf{50.9} & \textbf{37.2} & 24.6 & 38.3 & 17.0 & 34.8 &  \textbf{90.4} &  \textbf{75.5} &  97.3 &  67.9 &  \textbf{77.7} &  \textbf{84.8} \\
            \midrule
                \multirow{4}{*}{\texttt{Observation}} & \cellcolor{gray!30} 5 &  44.3  & 30.6  & 41.9  & 40.2  & \textbf{44.7}  & \textbf{41.4}  &   52.9 &  40.1 &  81.1 &  38.5 &  29.8 & 49.6 \\
                & \cellcolor{gray!45} 10 & 42.2  & 30.5  & \textbf{42.1}  & 41.9  & 36.2  & 38.8  &  57.4 &  53.1 &  \textbf{83.8} & 46.2 &  41.5 &  57.1\\
                & \cellcolor{gray!60} 25 & \textbf{44.6}  & 33.2  & 41.8  & 41.9  & 27.7  & 38.0  & \textbf{71.3} &  63.8 &  83.8 &  66.7 &  45.7 & 66.0 \\
                & \cellcolor{gray!75} 50 & 44.0  & \textbf{34.5}  & 41.1  & \textbf{41.9}  & 27.7  & 37.8  &  72.8 &  \textbf{73.2} &  83.8 &  \textbf{74.4} &  \textbf{56.4} &  \textbf{71.1}\\
            \midrule
                \multirow{4}{*}{\texttt{Summary}} & \cellcolor{gray!30} 2 &  34.6  & 15.7  & 26.9  & 26.5  & 36.2  & 29.9  &   68.4 &  39.6 &  56.8 & 50.0 & 73.4 &  61.5 \\
                & \cellcolor{gray!50} 5 & \textbf{36.6}  & \textbf{16.6} & \textbf{31.0} & \textbf{34.7}  & 38.3  & \textbf{32.5} &   81.6 &  57.0 &  70.3 &  60.3 &  86.2 &  75.1 \\
                & \cellcolor{gray!70} 10 & 34.5 & 14.7  & 29.3  & 31.6  & \textbf{40.4}  & 31.5  &  \textbf{93.4} &  \textbf{82.3} &  \textbf{91.9} &  \textbf{80.8} &  \textbf{94.7} &  \textbf{90.7} \\
            \bottomrule
        \end{tabular}
	}
 \vspace{-5pt}
        \caption{\textbf{Question answering performance} of RAG-based \texttt{GPT-3.5-turbo-16k}.  Optimal performance is in \textbf{bold}. Results are based on F1-score metric for answer prediction and recall@$k$ for recall accuracy; higher is better.}
	\label{tab:qa_rag_results}
 \vspace{-10pt}
\end{table*}

\section{Experimental Results}

We evaluate and analyze the comprehensive performance of all baseline methods for question answering (\Sref{ssec:qa-results}), event graph summarization (\Sref{ssec:event-results}), and multi-modal dialogue generation (\Sref{ssec:dialog-results}).

\subsection{Question Answering Task}
\label{ssec:qa-results}

Tables~\ref{tab:qa_results} and~\ref{tab:qa_rag_results} present the performance results for the question answering task. We find that:
\textbf{(1) LLMs with limited context length face challenges in understanding extremely long conversations} due to truncated context windows. Despite \texttt{gpt-4-turbo} emerging as the top-performing model with an overall score of 32.4, it notably lags behind the human benchmark of 87.9;
\textbf{(2) long-context LLMs can comprehend longer narratives, yet they are prone to generating hallucinations}. \texttt{gpt-3.5-turbo-16k} outperforms other approaches, but its performance on adversarial questions drops to a mere 2.1\%, as compared to 22.1\% using \texttt{Llama-2-Chat} and 70.2\% using \texttt{GPT-4-turbo} with 4K context windows. This indicates that LLMs can be easily misled into generating hallucinations when they are subjected to long contexts;
\textbf{(3) RAG is effective when conversations are stored as observations}. There is a noticeable 5\% improvement with \texttt{gpt-3.5-turbo} when the input is top 5 relevant observations instead of pure conversation logs. This improvement falters with an increase in the number of retrieved observations, suggesting that it is important to reduce the signal-to-noise (SNR) ratio in retrieved contexts for models to utilize the context accurately. Conversely, using session summaries as context does not significantly improve the performance despite high recall accuracies\footnote{For summary-based RAG models, the recall accuracy is based on retrieving the summary of the relevant session(s).}, likely due to loss of information during the conversion of dialogs to summaries.

The interesting finding is that \textbf{time reasoning and open-domain knowledge questions are the most challenging scenarios}.

(1) LLMs face challenges in understanding time concepts within dialogues, which is consistent with findings from other single-turn-based benchmarks focused on temporal reasoning capabilities for LLMs~\cite{wang2023tram}.

(2) LLMs struggle with open-domain knowledge and degrade in the RAG setting. This suggests that while certain open-domain knowledge may be embedded within the model's parameters, introducing improper context from inaccurate retrieval can lead to a decline in performance~\cite{mallen-etal-2023-trust}.

\begin{table*}[t!]
	\centering
	\small
	\resizebox{0.99\textwidth}{!}{
		\begin{tabular}{cccccccccc}
            \toprule
            \multirow{2}{*}{\textbf{Category}} & \multirow{2}{*}{\textbf{Model}} & \multirow{2}{*}{\textbf{Context Length}} & \multicolumn{3}{c}{\textbf{ROGUE}} & \multicolumn{3}{c}{\textbf{FactScore}} \\
            \cmidrule(lr){4-6} \cmidrule(lr){7-9} & & & \textbf{ROGUE-1} & \textbf{ROGUE-2} & \textbf{ROGUE-L} & \textbf{Precision} & \textbf{Recall} & \textbf{F1} \\
            \midrule
                \multirow{4}{*}{Base} & \texttt{Mistral-Instruct-7B} & 8K & 29.4 & 7.2 & 14.1 & 27.1 & 19.8 & 23.0 \\
                & \texttt{Llama-2-Chat-70B} & 4,096 & 28.1 & 9.3 & 14.8 & 36.3 & 22.7 & 28.3 \\
                & \texttt{GPT-4-turbo} & 4,096 & 38.8 & 11.4 & 20.6 & \textbf{51.6} & 41.8 & 45.1 \\
                & \texttt{GPT-3.5-turbo} & 4,096 & \textbf{41.1} & \textbf{13.5} & \textbf{20.9} & 45.3 & \textbf{46.5} & \textbf{45.9} \\
            \midrule
                Long context & \texttt{GPT-3.5-turbo-16K} & 16K & 36.2 & 8.5 & 16.4 & 42.3 & 37.8 & 39.9  \\
            \bottomrule
        \end{tabular}
	}
 \vspace{-1pt}
        \caption{\textbf{Event summarization performance} of \texttt{Base} and \texttt{Long-context} models. The optimal performance is shown in \textbf{bold}. Results are based on ROUGE and FactScore \cite{min2023factscore} metrics; higher is better.}
	\label{tab:summ_results}
\end{table*}

\begin{figure*}[t]
    \centering
    \includegraphics[width=0.99\textwidth]{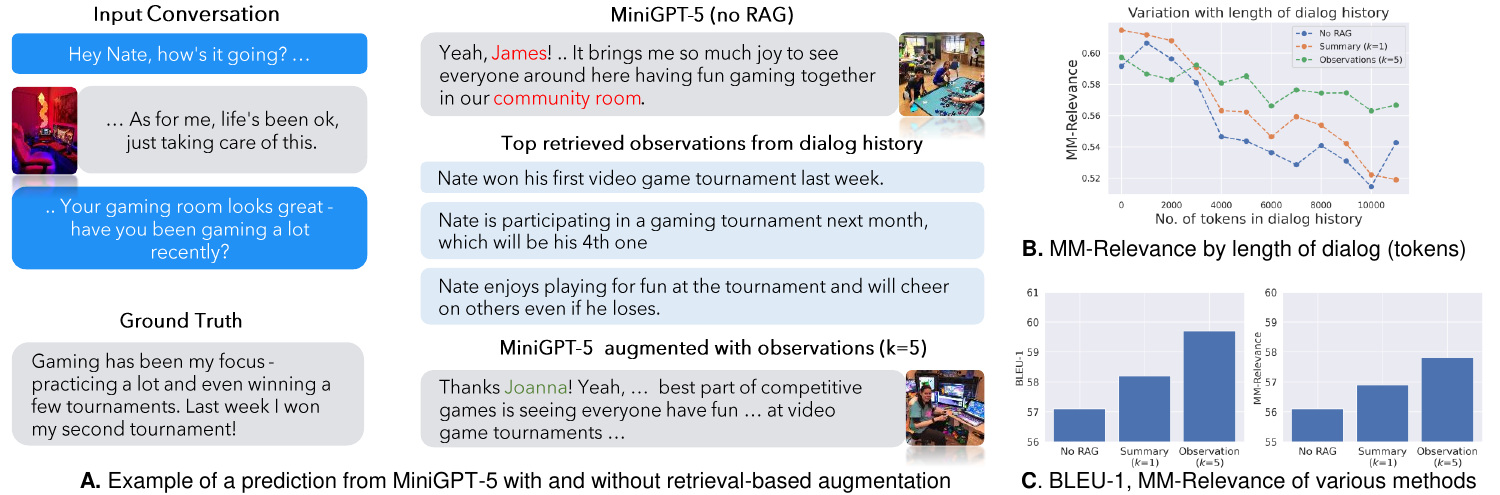}
    \vspace{-1pt}
    \caption{\textbf{Multimodal dialog generation performance of MiniGPT-5}. (A) an example of multimodal dialog predicted using MiniGPT5 with and without observation as retrieved context, (B) Variation of MM-Relevance score with length of dialog history, and (C) comparison of RAG-based MiniGPT-5 methods.\label{fig:minigpt5}}
    \vspace{-1pt}
\end{figure*}

\subsection{Event Summarization Task}
\label{ssec:event-results}

Table~\ref{tab:summ_results} presents results for the event summarization task. The use of incremental summarization with \textbf{\texttt{gpt-3.5-turbo} leads to the highest performance} in both recall and F1 score. While \texttt{gpt-4-turbo} records a 5.3\% improvement in precision over with \texttt{gpt-3.5-turbo}, it does not fare as well in terms of recall. The event summarization task requires long-range dependency to understand the temporal and causal connections between the events discussed by the speaker in multiple sessions (see Figure~\ref{fig:temporal-event}). Contrary to expectations, the \textbf{long-context model does not surpass the base model}, despite its capability for extended-range reasoning facilitated by a larger context window. \texttt{gpt-3.5-turbo-16k} exhibits a decline in both precision (by 3.0\%) and recall (by 8.7\%) compared to \texttt{gpt-3.5-turbo} which has a 4K context window. This suggests that \textbf{long-context models may not be proficient at utilizing their context appropriately}, which also aligns with similar findings in \citet{li2023loogle} as well as the QA task in \dataset{}. In terms of both the ROUGE and FactScore metrics, commercial models (\texttt{gpt-4-turbo}, \texttt{gpt-3.5-turbo}) significantly outshine their open-source counterparts. Nonetheless, there remains considerable scope for improving performance on this task.

From a manual analysis of predicted summaries, we identify five broad categories of event summarization errors made by LLMs: (1) \textbf{missing information} in events because the model fails to make temporal and/or causal connections over a lengthy conversation; (2) \textbf{hallucinations} i.e., models pad extra details that are either not present in the conversation or are part of a different event in the same session; (3) errors from \textbf{misunderstanding of dialog cues} such as humor or sarcasm is a distinctive issue with comprehension of dialogs; (4) inaccurate \textbf{speaker attributions}; and (5) insignificant dialogs that are wrongly considered as \textbf{salient} events. See examples in Table~\ref{tab:summary_errors} in the Appendix.

\subsection{Multi-Modal Dialog Generation Task}
\label{ssec:dialog-results}

Figure~\ref{fig:minigpt5} illustrates the effectiveness of various MiniGPT-5 training variants in multi-modal dialogue generation. Incorporating context into training enhances performance, with the inclusion of observation as context yielding significantly improved results. For instance, in Figure~\ref{fig:minigpt5}A, the retrieved observations contain information about the speaker's experience in video game tournaments, which leads to the prediction of dialog and images that are more faithful to the speaker's persona.
This observation is consistent with earlier findings from the QA task as well (see Table~\ref{tab:qa_rag_results}). Also, we observe that the MM-Relevance score drops with an increase in the length of dialog history (see Figure~\ref{fig:minigpt5}B). Retrieval-augmented generation alleviates the drop in MM-Relevance to some extent. 

\section{Conclusion}
 We develop a human-machine pipeline to collect \dataset{}, a datset of 50 high-quality very long conversations, each encompassing 300 turns and 9K tokens on avg., over up to 35 sessions, and propose an evaluation framework consisting of three tasks that evaluate models' proficiency in long conversations. Our experiments show that LLMs struggle to comprehend long-term narratives within the dialog and fail to draw temporal and causal connections between events discussed by speakers.
\section{Limitations}

\paragraph{Hybrid human-machine generated data.} Our dataset is sourced primarily from text generated by LLMs. We pursued this method, which has quickly emerged as a popular alternative to time-intensive manual data collection \cite{kim-etal-2023-soda, jang2023conversation}, to avoid the logistical and legal complexities of collecting very long-term real-world conversations at scale. We ensure that the dataset mirrors real-world interactions as much as possible by having human annotators verify and edit the generated conversations. However, we acknowledge that this dataset may not fully reflect the nuances of real-world online conversations. 

\paragraph{Limited exploration of multimodal behavior.} Since the images in our dataset are sourced from the web, they do not demonstrate the visual long-term consistencies that are usually exhibited in personal photos (e.g., appearance, home environment, people and pets, etc.). Consequently, we find that the images in our dataset can be replaced with their captions without much loss of information, except for cases where OCR is required. Nevertheless, our work is a first step toward research into the multimodal aspect of very long-term conversations.

\paragraph{Language.} Our LLM-based pipeline for generating long-term conversations has been developed for the English language only. However, our pipeline can be made to work with any other language using an LLM that is proficient at that language and appropriate translations of our prompts.

\paragraph{Closed-source LLMs.} We use state-of-the-art LLMs in our dialog generation pipeline to create a dialog dataset that is as realistic as possible. Unfortunately, this meant employing the strongest commercial LLMs available through a paid API, similar to many concurrent works that generate synthetic conversations \cite{zhong2023memorybank, lu2023memochat}. We will make the code for our generative pipeline publicly available in the hope that it can be made to work effectively with state-of-the-art open-source LLMs in the future.

\paragraph{Evaluation of long-form NLG.} LLMs are prone to generating verbose answers even when prompted to answer in short phrases. This creates challenges in evaluating the correctness of answers provided by LLMs and has been widely documented in NLP literature \cite{chang2023booookscore, xu2023critical, krishna2023longeval}. Our evaluation framework suffers from the same challenges when used for experimenting with LLMs.

\section{Broader Impacts}
We adopt and improve a framework of generative agents introduced in \citet{park2023generative} for the generation of long-term conversations. Consequently, the ethical concerns of generative agents outlined by \citet{park2023generative} apply to our work as well, especially since the goal of our framework is to make the conversations as realistic as possible.

Specifically, conversational agents that can pose as human beings with a realistic life, as enabled by the temporal event graphs in our framework, pose the risk that users may form parasocial relationships with such agents that may affect their lives adversely. We recommend that any practical deployment of the generative frameworks mentioned in our work be always prefaced with a disclaimer about the source of the dialogs.

Second, the use of multimodal LLMs \cite{zheng2023minigpt} to generate images conditioned on dialog can lead to the propagation of misinformation and social biases, especially if the conversational agent can be coerced into parroting false information or dangerous opinions.

Third, it is tempting to use generative agents to substitute real humans for a process, especially when there are significant challenges in working with humans for a particular goal e.g., collecting real-world interactions between humans over a year or more. Care must be taken to ensure that such substitutes are not made in studies whose outcomes may be used to make real-world decisions with tangible impacts on humans. Our work is merely a study of model comprehension in very long-term conversations. We do not make any recommendations for real-world policies based on this study and advise potential users of our framework to avoid making such recommendations as well.

\bibliography{custom}


\appendix
\section*{Appendix Overview}
\noindent The appendix is organized as follows:\\
\textbf{Section A}: Details of generative pipeline for the \dataset{} dataset.\\
\textbf{Section B}: Statistics of \dataset{} dataset, license for data release and annotator details.\\
\textbf{Section C}: Experimental setup and implementation details.\\
\textbf{Section D}: Additional results from evaluation on the \dataset{} benchmark.\\

\begin{figure*}[t]
    \centering
    \includegraphics[width=0.99\textwidth]{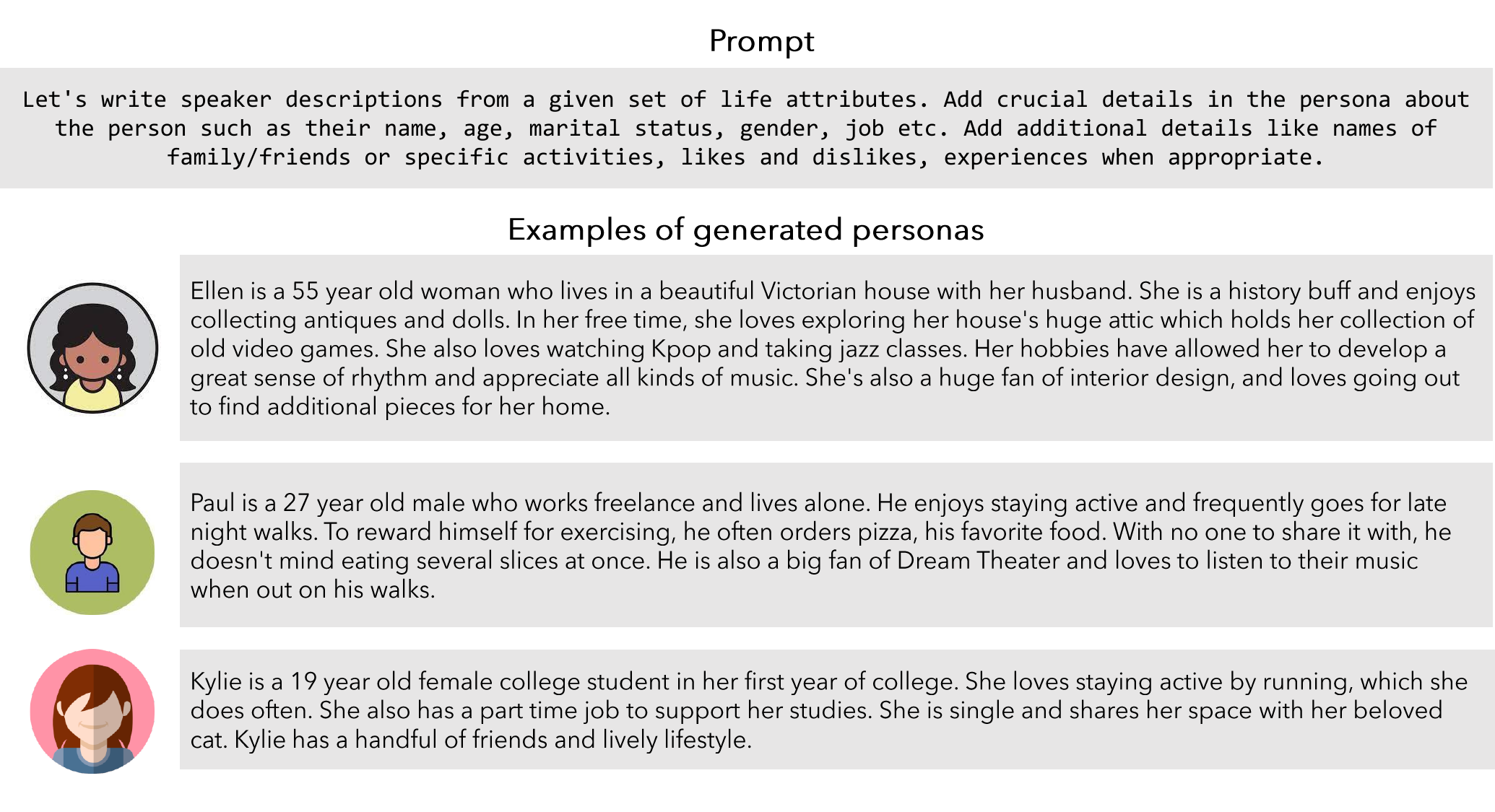}
    \vspace{-10pt}
    \caption{\textbf{Prompt for persona statement ($p$) generation and examples of personas in \dataset{}.} The prompt used to generate expanded persona statements ($p$) from initial personas ($p_c$) for the virtual agents in our conversation generation pipeline (top) and select examples of persona statements present in the \dataset{} dataset.\label{fig:persona}}
\end{figure*}

\section{Generative Pipeline for \dataset{}}
\label{sec:generative-pipeline-appendix}

\subsection{Persona}
\label{ssec:persona_appendix}
We assign unique persona statement $p$ to each agent $\mathcal{L}_i$. For this, we select a range of initial persona statements $p_c$ from the MSC dataset~\cite{xu2022beyond}, each encompassing 4 to 5 sentences. We employ \texttt{gpt-3.5-turbo} as $\mathcal{M}$ to expand these into full persona statement $p$, conditioning $\mathcal{M}$ on the chosen statements $p_c$. The prompt used for converting a short list of speaker attributes from the MSC dataset \cite{xu2022beyond} into a complete persona summary is presented in Fig.~\ref{fig:persona}. We also use a single example of speaker attribute $\rightarrow$ persona summary as an in-context demonstration along with the prompt. A small selection of personas showcasing the diversity of speakers in the \dataset{} dataset is demonstrated in Fig.~\ref{fig:persona}.

\subsection{Temporal Event Graph}
\label{ssec:event_appendix}

\begin{figure*}[t]
    \centering
    \includegraphics[width=0.99\textwidth]{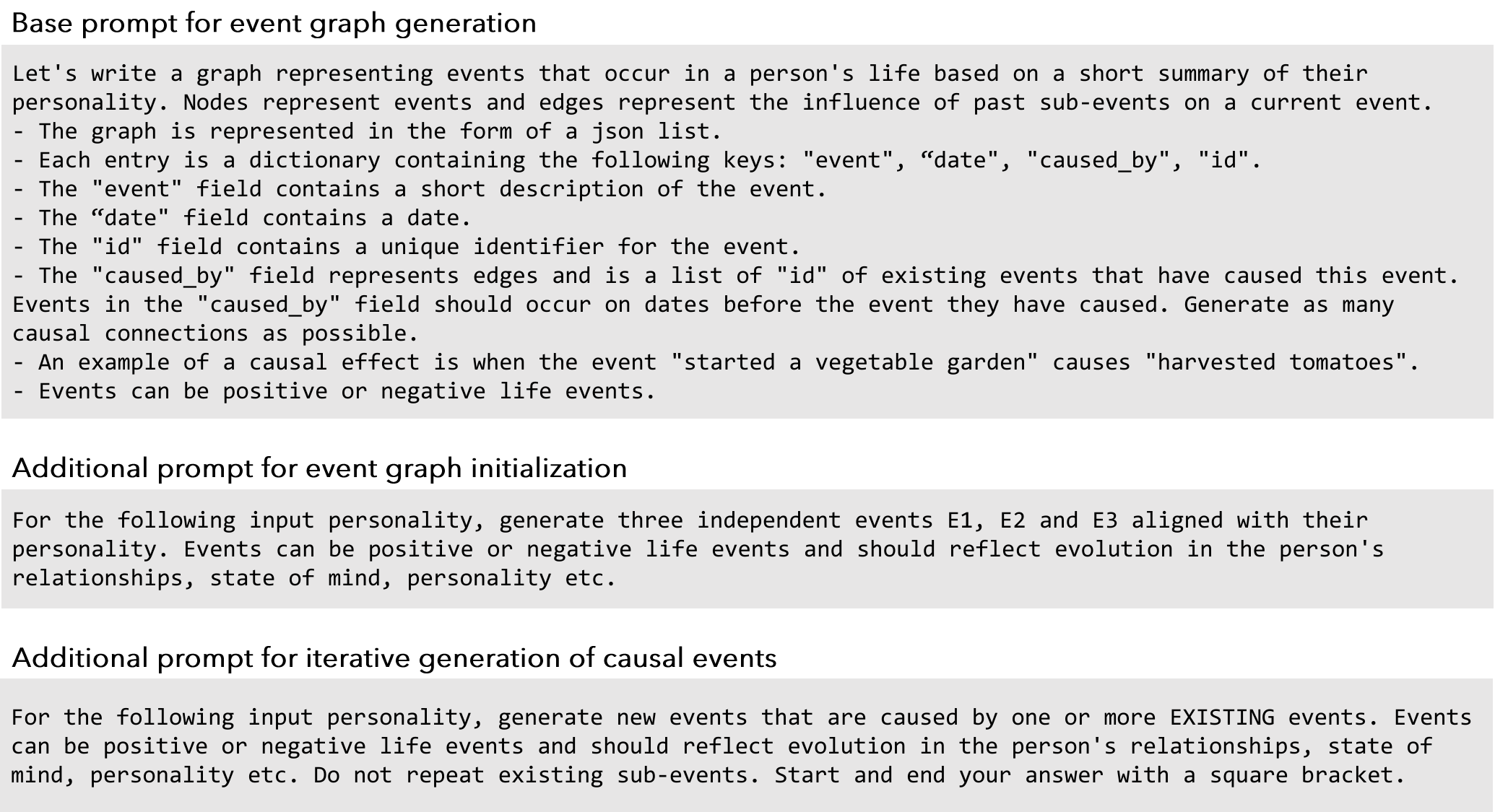}
    \caption{\textbf{Prompts for temporal event graph generation.} The prompt used to generate complete personas for the LLMs in our conversation generation pipeline (top) and examples of personas present in the \dataset{} dataset.\label{fig:event_prompt}}
\end{figure*}

\begin{figure*}[t]
    \centering
    \includegraphics[width=0.99\textwidth]{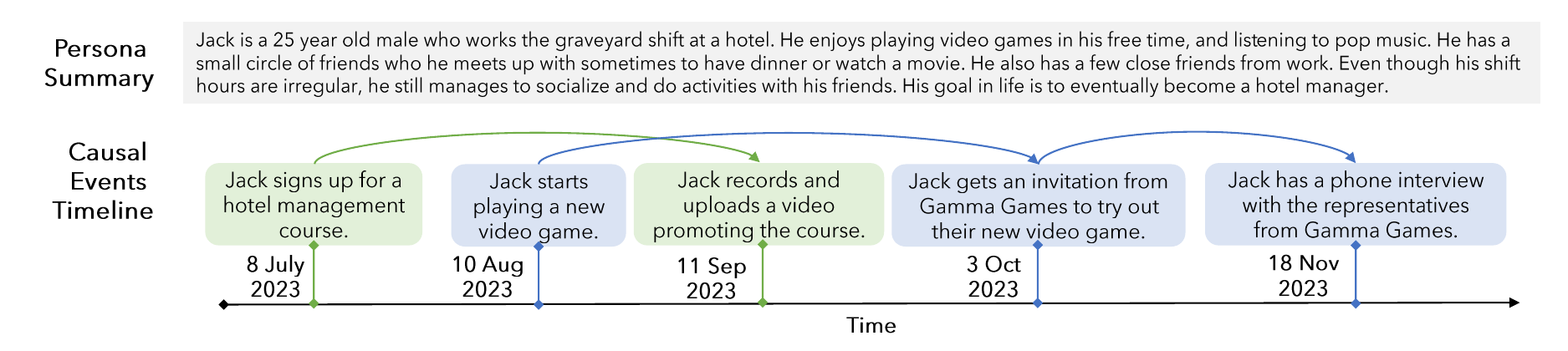}
    \vspace{-0.5cm}
    \caption{\textbf{Temporal Event Graph $\mathcal{G}$ Creation.} Each event is generated in accordance with the specified persona $p$ and causal connections $l$ between events are depicted to illustrate the casual relationships among them.}
    \label{fig:temporal-event}
\end{figure*}

As outlined in Sec.~\ref{ssec:temporal-event}, we use an iterative process for generating event graphs consisting of causally connected events based on a given persona summary. The base prompt for describing the constitution of the event graph, the nature of events and causal connections between events is shown in Fig.~\ref{fig:event_prompt}. First, the base prompt is used along with the prompt for event graph initialization to generate three independent events relevant to a given personality. Then, the base prompt is combined with the prompt for the iterative generation of events to continue generating events that are caused by one or more of the events that are already present in the graph. See an example of a persona and the corresponding temporal event graph in Fig.~\ref{fig:temporal-event}. In the example, Jack aspires to be a hotel manager. Consequently, he enrolls in a hotel management course in July, and after three months, he expresses his excitement about the course on social media. In a similar vein, his passion for gaming results in an invitation from a well-known gaming company.

\subsubsection{Virtual Agent Architecture}
\label{sssec:agent_appendix}

As outlined in Section~\ref{ssec:llm-agent}, the virtual agents in our generative pipelines are composed of two mechanisms, \textit{Reflect \& respond} \cite{park2023generative} and \textit{Image sharing \& response}.

\begin{figure*}[t]
    \centering
    \includegraphics[width=0.99\textwidth]{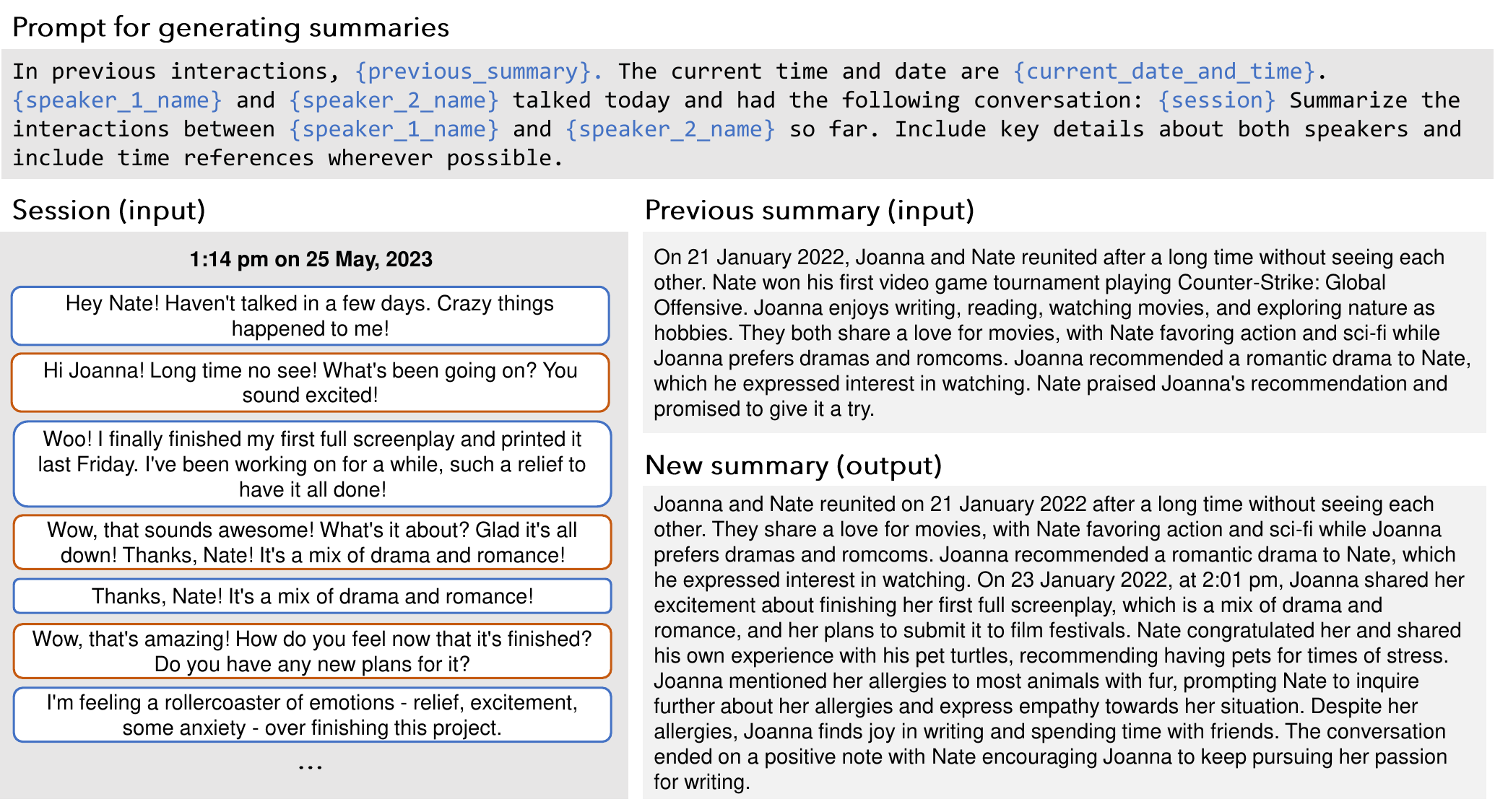}
    \caption{\textbf{Prompt for generating conversation summaries.} The prompt used to iteratively generate a summary for the current session by conditioning on summary from preceding sessions and the raw conversation logs of the current session (top); and an example of inputs for the prompt and corresponding output summary of a session from the \dataset{} dataset. \label{fig:summary_prompt}}
\end{figure*}

\begin{figure*}[t]
    \centering
    \includegraphics[width=0.99\textwidth]{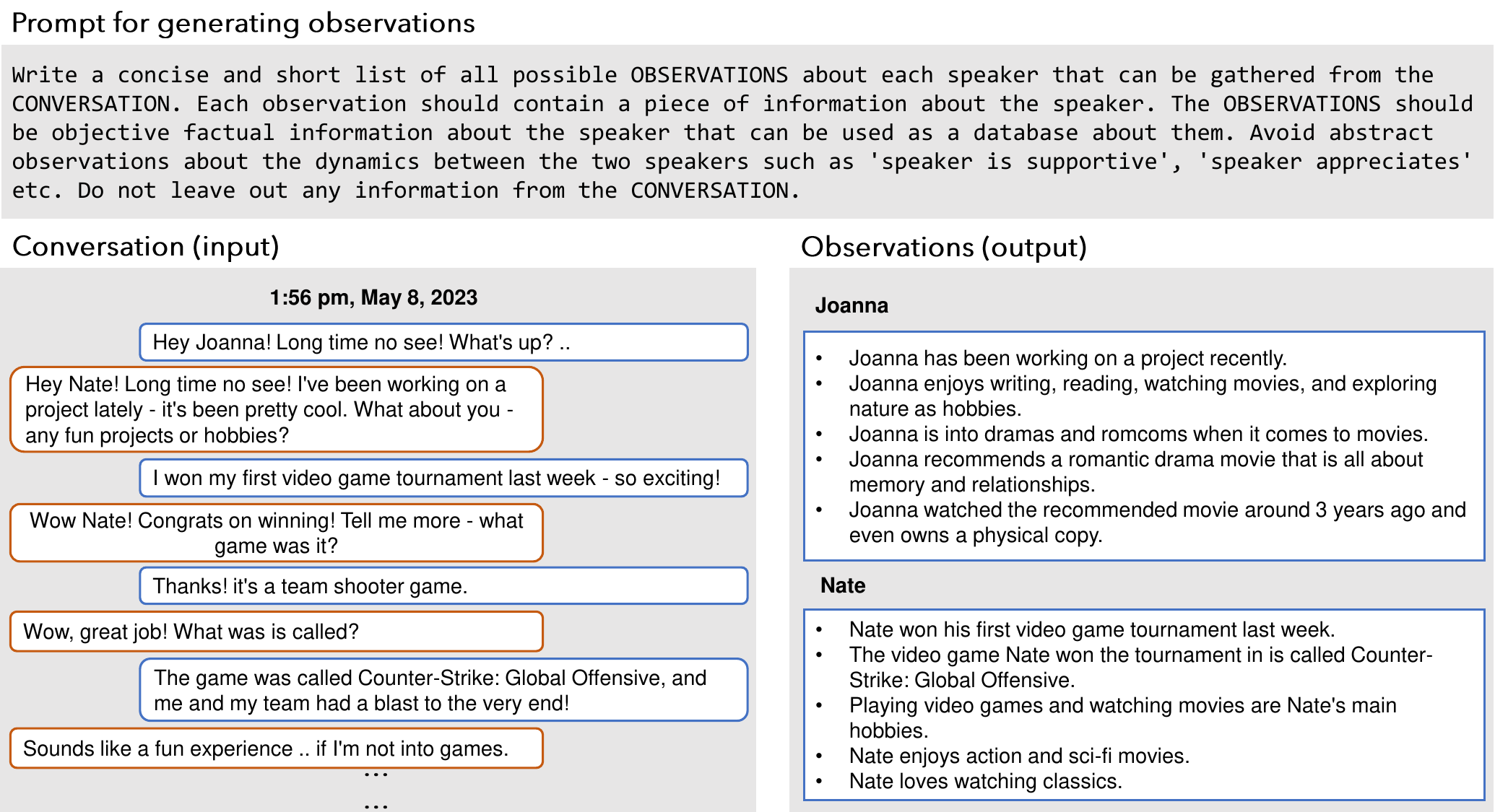}
    \caption{\textbf{Prompts for generating observations from conversations.} The prompt used to generate observations from a conversation (top); and an example of inputs for the prompt and corresponding output observations for a session from the \dataset{} dataset.\label{fig:observation_prompt}}
\end{figure*}

\paragraph{Reflect \& respond.} This mechanism operates over a combination of short-term and long-term memory. The short-term memory is a summary of a session that is conditioned on the summary from a previous session. See the prompt given to LLMs in our pipeline for generating summaries, and an example of a generated summary, in Fig.~\ref{fig:summary_prompt}. The long-term memory is a database of \textit{observations} about each speaker, that are essentially assertive statements about the speaker's persona and life. See the prompt given to LLMs in our pipeline for generating observations, and an example of observations extracted from a conversation, in Fig.~\ref{fig:observation_prompt}. In practice, the conversation is annotated with turn IDs for each turn, and the model is also instructed to indicate the turn IDs that directly contribute to each observation. This allows us to keep track of the evidence when using observations as the context for RAG-based models used in our experiments (see Section~\ref{sec:baselines}).

\begin{figure*}[t]
    \centering
    \includegraphics[width=0.99\textwidth]{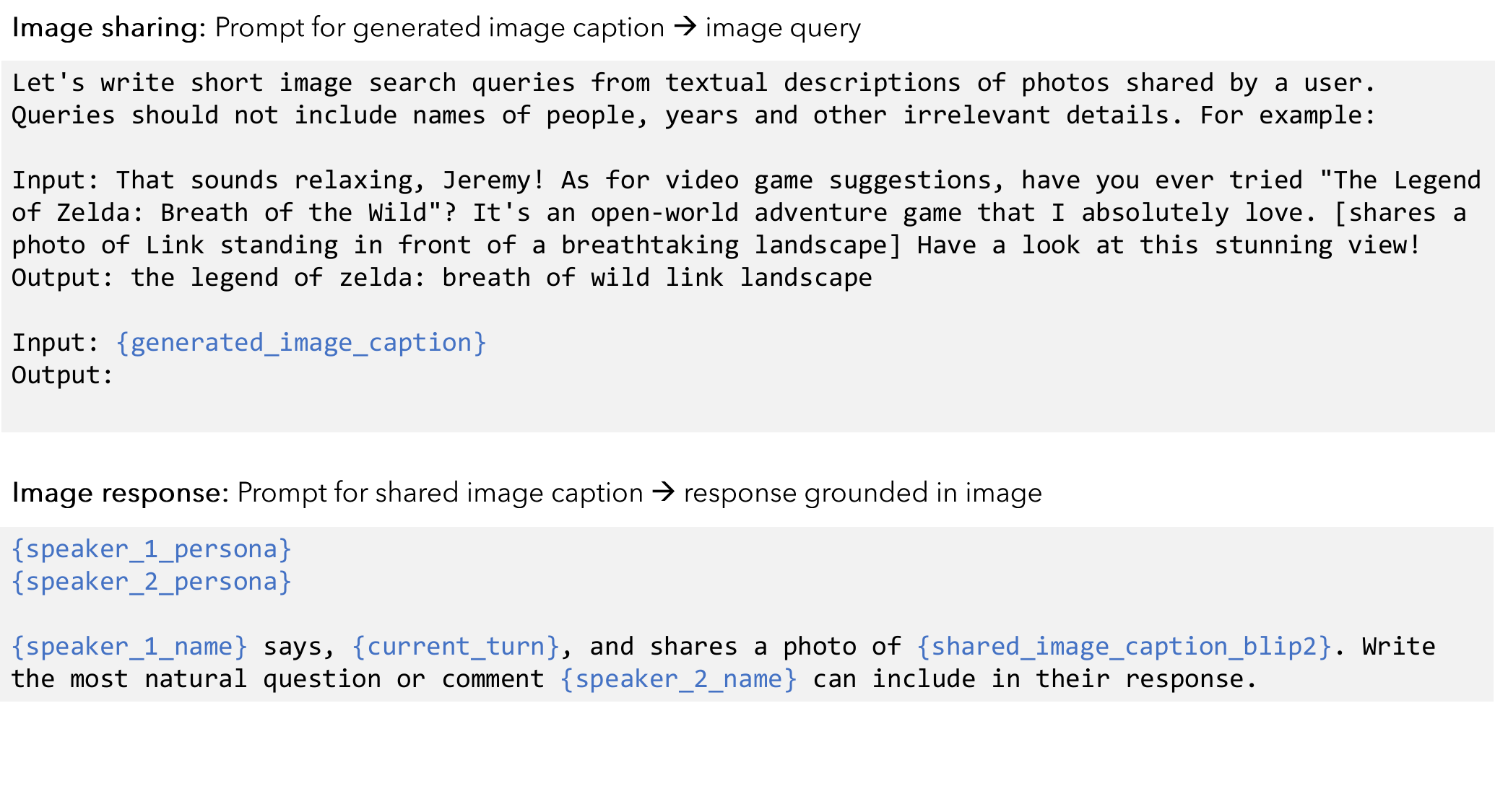}
    \vspace{-30pt}
    \caption{\textbf{Prompts for image-sharing and image-response behavior.} The prompt used to convert a caption generated by the virtual agent into an image query for the web-based image crawler in our pipeline (top), and the prompt used to generate a response grounded in the image shared by a virtual agent during a conversation as well as the personas of the respective speakers (bottom).\label{fig:image_prompt}}
\end{figure*}

\paragraph{Image sharing \& response.} See prompts for implementing image-sharing and image-response behaviors in Figure~\ref{fig:image_prompt}. 

\subsection{Human Filtering}
\label{ssec:edits_appendix}

\begin{figure*}[t]
    \centering
    \includegraphics[width=0.99\textwidth]{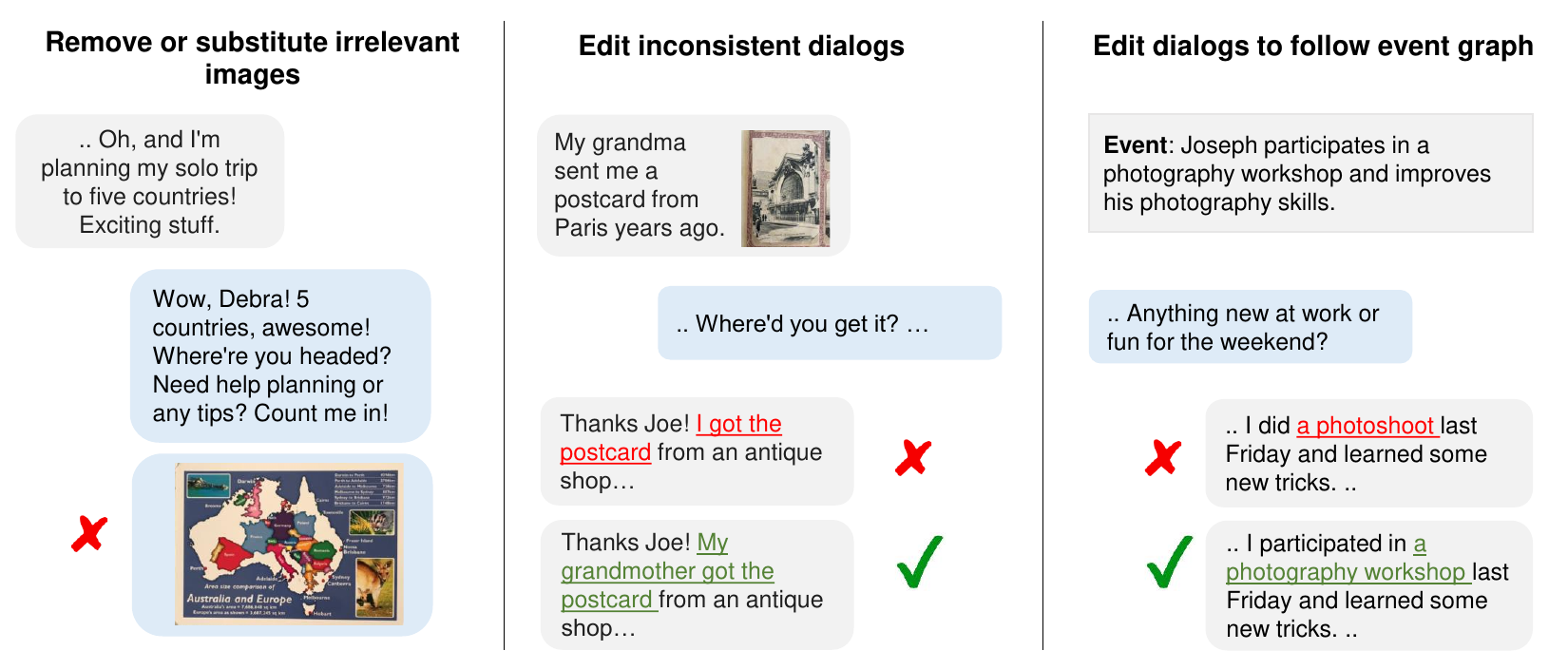}
    \caption{\textbf{Example of edits made by annotators.} Human annotators are instructed to make edits in the LLM-generated conversations to remove irrelevant The prompt used to generate complete personas for the LLMs in our conversation generation pipeline (top) and examples of personas present in the \dataset{} dataset.\label{fig:conv_edits}}
\end{figure*}

Human annotators are instructed to edit the LLM-generated conversations in the following scenarios:
\begin{itemize}
    \item Remove an image if it is not relevant to the current dialog or the conversation.
    \item Add context about an image to the current speaker's dialog if it is not discussed by them but the subsequent speaker has reacted to the image.
    \item Replace an image if it does not match the caption that was used to query for images.
    \item Edit the dialog when the information present in the dialog is inconsistent with something said (or shared through an image) in earlier or later turns.
    \item Edit the dialog to ensure that the details in the conversation are consistent with those given in the event for the session.
    \item Remove any events from the event graph if they do not appear in the conversation.
\end{itemize}

See an example of some edits in Fig.~\ref{fig:conv_edits}.

\section{Dataset}

\subsection{Dataset Statistics}
See a breakdown of the statistics of the conversations in the \dataset{} dataset in the top panel of Table~\ref{tab:dataset_statistics}. Also, see a breakdown of the statistics of the annotations in the evaluation benchmark in the bottom panel of Table~\ref{tab:dataset_statistics}.

\begin{table}[t]
	\centering
	\small
	\resizebox{\linewidth}{!}{
		\begin{tabular}{lc}
            \toprule
            \rowcolor{gray!20} Conversation Statistics & \# Counts \\
            \midrule
             Total. \# conversations $h$. & 50 \\
             Avg. \# sessions $k$. in conversation $h$ & 19.3 \\
             Avg. \# turns $j$. in session $k$ & 15.8 \\
            \midrule
             Avg. \# tokens. conversation $h$ & 9,209.2 \\
             Avg. \# tokens. dialogue $h_{k_j}$ of turn $j$ in session $k$ &  30.2 \\
             Avg. \# tokens. observation $o_{k_j}$ of turn $j$ in session $k$ & 18.2 \\
             Avg. \# tokens. summary $w_k$ of session $k$ & 127.4 \\
             \midrule
             \rowcolor{gray!20} QA Benchmark Statistics & \\
             \midrule
             \# questions. single-hop retrieval & 2,705 (36\%)\\
             \# questions. multi-hop retrieval & 1,104 (14.6\%)\\
             \# questions. temporal reasoning &  1,547 (20.6\%)\\
             \# questions. open domain knowledge & 285 (3.9\%)\\
             \# questions. adversarial & 1,871 (24.9\%)\\
             \textbf{Total. \# questions.} & \textbf{7,512} \\
             \midrule
             \rowcolor{gray!20} Event Summarization Statistics & \\
             \midrule
             Avg. \# ground truth events. in conversation $h$ & 24.2 \\
             Avg. \# tokens. event summary & 896.5 \\
             \midrule
             \rowcolor{gray!20} Multi-modal Dialogue Generation Statistics & \\
             \midrule
             Avg. \# images. in conversation $h$ & 32.3 \\
            \bottomrule
        \end{tabular}
	}
        \caption{\textbf{Dataset Statistics} of conversation and corresponding benchmark}
	\label{tab:dataset_statistics}
\end{table}

\subsection{Dataset License}
The \dataset{} dataset will be released under the CC BY-NC 4.0 DEED license.\footnote{\url{https://creativecommons.org/licenses/by-nc/4.0/}}

\subsection{Annotator Details}
The annotators who worked on the \dataset{} dataset were in-house annotators and we were unable to obtain their demographics due to the confidential nature of such information.

\section{Experimental Setup}
\label{sec:experiments_appendix}

\subsection{Baselines}
\label{ssec:baselines-appendix}
The conversations in the \dataset{} dataset are composed of natural language dialogs and images that require higher-order reasoning and multimodal coreference resolution, respectively. From initial studies, we observed that multimodal coreference resolution can be performed effectively by replacing images in \dataset{} with their captions generated using BLIP-2 \cite{li2023blip}, and using state-of-art LLMs to reason over natural language text interleaved with image captions. Hence, our experiments for the question answering and event summarization tasks are conducted using LLMs. We use the images directly only for experiments on the multimodal dialog generation task.

\paragraph{Question Answering.}
We carry out experiments using three distinct methodologies:
(1) \textbf{Base} involves utilizing LLMs to directly conduct the task within a constrained context.
The task description comes after the dialogue history. 
To accommodate the restricted context window size, earlier dialogues are omitted; 
(2) \textbf{Long-context} employs LLMs with an extended context window to expose the models to as much dialogue context as possible;
(3) \textbf{Retrieval-augmented Generation (RAG)} involves retrieving relevant context from a database of dialog history, observations, or session-level summaries. \textit{Observations} are assertions about each speaker extracted from the dialog history as described in \Sref{ssec:llm-agent}, see an example in Figure~\ref{fig:observation_prompt}. Session-level \textit{summaries} are concise summaries of the conversation that takes place in each session, see an example in Figure~\ref{fig:summary_prompt}.

For the retrieval model, we employ DRAGON~\cite{lin-etal-2023-train}.
In the \textit{Base}, we utilize Mistral-7B~\cite{jiang2023mistral}, LLama-70B-chat~\cite{touvron2023llama}, \texttt{gpt-3.5-turbo}~\footnote{https://platform.openai.com/docs/models/gpt-3-5}, and \texttt{gpt-4-turbo}~\footnote{https://platform.openai.com/docs/models/gpt-4-and-gpt-4-turbo}. 
To assess the effectiveness in practical scenarios for \textit{Long-context} and \textit{RAG}, we draw comparisons using variants of \texttt{gpt-3.5-turbo}.
We do not report the performance of long-context fine-tuned open-source models~\cite{chen2023longlora} or those utilizing sliding window~\cite{bertsch2024unlimiformer, dao2022flashattention} due to the variability inherent across different open-source models and the potential reduction in their capability on shorter context.

\paragraph{Event Summarization.}
We present experiments conducted in two distinct configurations.
We use both the \textbf{Base} and \textbf{Long-context} setups from the question answering task, but we refrained from including RAG since summarization requires a comprehensive understanding of the entire dialogue, rather than just retrieving a specific portion. A notable distinction in our approach, compared to the question-answering task, lies in our handling of the context. 
Specifically, we employ an iterative process of creating a summary of a preceding session and then use that summary as a basis to generate the summary for the subsequent session \cite{chang2023booookscore}. Further, we use a single in-context demonstration of input and output to guide the model toward selecting only significant life events for the summary.

\paragraph{Multi-modal Dialogue Generation.}
For evaluating multi-modal dialogue generation, we train MiniGPT-5~\cite{zheng2023minigpt} on 50 conversations generated using our automated pipeline (without human filtering) as detailed in \Sref{sec:dataset}.
Three distinct versions of the model were developed, each with varying training data:
(1) \textbf{Base} trains on preceding dialogue turns;
(2) \textbf{+ summary} trains on both prior dialogue turns and a global summary of the ongoing conversation;
(3) \textbf{+ observation} trains on both preceding dialogue turns and relevant observations retrieved from the conversation history.
For each of these models, we started with a MiniGPT-5 checkpoint pretrained on the MMDialog dataset \cite{feng2023mmdialog}.

\subsection{Implementation Details}
We use OpenAI API and Huggingface~\cite{wolf-etal-2020-transformers}, as of January 2024, with specific settings of $temperature$ set to 0 and $top_p$ set to 1 for evaluation of the \dataset{} benchmark. All experiments, including those for RAG-based models, MiniGPT-5 training, and inference, are conducted on an Nvidia A6000 server with FP32. We report results from a single inference run for each model in our experiments. For MiniGPT-5, we used the hyperparameters recommended in the original codebase and trained our models for 10 epochs, which took approximately 30 hours on a single A6000 GPU.

We use the default implementations of BLEU\footnote{\url{https://www.nltk.org/_modules/nltk/translate/bleu_score.html}}, ROUGE\footnote{\url{https://pypi.org/project/rouge/}}, BertScore\footnote{\url{https://pypi.org/project/bert-score/}}, FactScore\footnote{\url{https://github.com/shmsw25/FActScore}} metrics in their respective Python packages in our evaluation protocol.

\section{Results}

\begin{table}[t!]
	\centering
	\small
	\resizebox{\linewidth}{!}{
		\begin{tabular}{ccccc}
            \toprule
             \textbf{Category} & \textbf{top-$k$} & \textbf{BLEU-1/2} & \textbf{Rouge-L} & \textbf{MM-R} \\
            \midrule
                \texttt{Base}  &  - & 57.1 / 34.2 & 12.4 & 56.1 \\
            \midrule
                \texttt{+ summary} & \cellcolor{gray!30} 1 & 58.2 / 34.1 & 12.8 & 56.9 \\
                \texttt{+ summary} & \cellcolor{gray!45} 2 & 56.5 / 32.8 & 12.1 & 55.1 \\
                \texttt{+ summary} & \cellcolor{gray!60} 5 & 56.1 / 32.5 & 12.0 & 55.2 \\
            \midrule
                \texttt{+ observation} & \cellcolor{gray!30} 5 & \textbf{59.7} / \textbf{35.1} & \textbf{13.6} & \textbf{57.8} \\
                \texttt{+ observation} & \cellcolor{gray!45} 10 & 59.1 / 34.9 & 12.8 & 57.1 \\
                \texttt{+ observation} & \cellcolor{gray!60} 25 & 58.5 / 34.2 & 12.0 & 56.5 \\
            \bottomrule
        \end{tabular}
	}
        \caption{\textbf{Multi-modal dialogue generation performance} comparison between different training variants of MiniGPT-5.
        The optimal performance is shown in \textbf{bold}.}
	\label{tab:dialog_results}
\end{table}

\subsection{Event Summarization Task}
\label{ssec:appendix-event-summ}

See an example of the five broad categories of event summarization errors made by LLMs, outlined in Section~\ref{ssec:event-results}, in Table~\ref{tab:summary_errors}.

\begin{table*}[t!]
	\centering
	\small
	\resizebox{\linewidth}{!}{
		\begin{tabular}{P{0.7in}>{\columncolor{blue!10}}P{2.2in}>{\raggedright}P{2.8in}P{1.8in}}
            \toprule
             \textbf{Error Type} & \textbf{Explanation} & \textbf{Ground truth event \textit{or} relevant dialogs} & \textbf{Predicted event} \\
            \midrule
            Missing information & Key details about event are omitted because the model fails to make causal and temporal connections over a long conversation. & Joanna submits her third screenplay on loss, identity, and connection to a film contest & Joanna submits her recent screenplay to a film contest. \\
            \midrule
            Hallucination & Non-existent details or details from a different event are padded onto an event & \textit{N}: `The gaming party was a great success!' \newline \textit{N}: `... said they'd want to do it again next month!' \newline \textit{N}: `On another note, I made vegan ice cream ...' & Nate's vegan ice cream is a huge success and people want to do it again next month. \\
            \midrule
            Misunder- -standing of dialog cues & e.g., model confuses a light-hearted statement from a speaker as a serious statement & \textit{J}: `.. these trails that made me feel like writing a drama.' \newline \textit{N}: `.. go together .. Maybe I'll start to think of a drama myself and write a screenplay ...' \newline \textit{J}: `Haha, now that would be something! ...' & Nate considers writing his own drama screenplay. \\
            \midrule
            Speaker attribution & Event is attributed to the wrong speaker & Nate invites Joanna to try his homemade lactose-free ice cream. & Joanna invites Nate to her home to try her dairy-free ice cream recipe. \\
            \midrule
            Saliency & Unimportant interactions in the conversation are considered significant by model & \textit{N}: Hey Joanna, what's been up since we last chatted? How's it going? & Nate asks Joanna how she has been she they last talked.\\
            \bottomrule
        \end{tabular}
	}
        \caption{\textbf{Taxonomy of errors in LLM-generated event summaries.} Five types of errors predominantly occur in the event summaries generated by LLMs. Examples are based on predictions from \texttt{gpt-3.5-turbo}.}
	\label{tab:summary_errors}
\end{table*}

\subsection{Multimodal Dialog Generation Task}

Results from evaluation of various version of MiniGPT-5 model on the multimodal dialog generation task in the \dataset{} benchmark is in Table~\ref{tab:dialog_results}.

\end{document}